\pdfoutput=1

\documentclass[sigconf]{acmart}

\usepackage{color,soul}
\usepackage{multirow}

\usepackage{latexsym}
\usepackage{makecell} 

\AtBeginDocument{%
  \providecommand\BibTeX{{%
    \normalfont B\kern-0.5em{\scshape i\kern-0.25em b}\kern-0.8em\TeX}}}

\setcopyright{acmcopyright}
\copyrightyear{2023} 
\acmYear{2023} 
\setcopyright{acmlicensed}\acmConference[WWW '23]{Proceedings of the ACM Web Conference 2023}{May 1--5, 2023}{Austin, TX, USA}
\acmBooktitle{Proceedings of the ACM Web Conference 2023 (WWW '23), May 1--5, 2023, Austin, TX, USA}
\acmPrice{15.00}
\acmDOI{10.1145/3543507.3583236}
\acmISBN{978-1-4503-9416-1/23/04}

\author{Kunpeng GUO}
\authornote{Corresponding Author}
\orcid{0000-0002-0692-0057}
\affiliation{%
    \institution{The QA Company SAS}
    \institution{Université Jean Monnet}
    \city{Saint Etienne}
    \country{France}
}
\email{kunpeng.guo@univ-st-etienne.fr}

\author{Dennis Diefenbach}
\orcid{0000-0002-0046-2219}
\affiliation{%
    \institution{The QA Company SAS}
    \institution{Université Jean Monnet}
    \city{Saint Etienne}
    \country{France}
}
\email{dennis.diefenbach@the-qa-company.com}

\author{Antoine Gourru}
\orcid{0000-0003-3571-2430}
\affiliation{%
    \institution{Université Jean Monnet}
    \city{Saint Etienne}
    \country{France}
}
\email{antoine.gourru@univ-st-etienne.fr}

\author{Christophe Gravier}
\orcid{0000-0001-8586-6302}
\affiliation{%
    \institution{Université Jean Monnet}
    \city{Saint Etienne}
    \country{France}
}
\email{christophe.gravier@univ-st-etienne.fr}

\renewcommand{\shortauthors}{Kunpeng GUO, et al.}

\begin{document}
\newcommand{\frameworkName}{WebExtractor}

\ccsdesc[500]{Information systems~Data extraction and integration}

\keywords{Knowledge Graph Completion, Wikidata, Question Answering, Web Extraction, Linking, Web Crawling, Web Scraping}

\title{Wikidata as a seed for Web Extraction}

\begin{abstract}
Wikidata has grown to a knowledge graph with an impressive size. To date, it contains more than 17 billion triples collecting information about people, places, films, stars, publications, proteins, and many more.
On the other side, most of the information on the Web is not published in highly structured data repositories like Wikidata, but rather as unstructured and semi-structured content, more concretely in HTML pages containing text and tables. Finding, monitoring, and organizing this data in a knowledge graph is requiring considerable work from human editors. The volume and complexity of the data make this task difficult and time-consuming.
In this work, we present a framework that is able to identify and extract new facts that are published under multiple Web domains so that they can be proposed for validation by Wikidata editors. The framework is relying on question-answering technologies. We take inspiration from ideas that are used to extract facts from textual collections and adapt them to extract facts from Web pages. For achieving this, we demonstrate that language models can be adapted to extract facts not only from textual collections but also from Web pages.
By exploiting the information already contained in Wikidata the proposed framework can be trained without the need for any additional learning signals and can extract new facts for a wide range of properties and domains. Following this path, Wikidata can be used as a seed to extract facts on the Web.
Our experiments show that we can achieve a mean performance of 84.07 at F1-score. Moreover, our estimations show that we can potentially extract millions of facts that can be proposed for human validation. The goal is to help editors in their daily tasks and contribute to the completion of the Wikidata knowledge graph.
\end{abstract}

\maketitle


\section{Introduction}
Knowledge Graphs (KGs) are data structures that allow us to organize and aggregate structured information about heterogeneous entities. Known examples of such knowledge graphs are Freebase~\cite{bollacker.2008}, NELL~\cite{carlson2010toward}, DBpedia~\cite{auer.2007}, Yago~\cite{Mahdisoltani2014YAGO3AK} and more recently Wikidata~\cite{vrandevcic2014wikidata}.\\
While knowledge graphs like Wikidata contain an incredible amount of information, most of the knowledge on the Web is published on HTML websites in structured, semi-structured, and unstructured content. Aggregating this information in knowledge graphs is beneficial since one can interlink Web resources, and further allow the machine to exploit them for Natural Language Processing (NLP) tasks such as entity linking~\cite{mendes2011dbpedia}, query expansion~\cite{dalton2014entity}, co-reference resolution~\cite{rahman2011coreference}, and question
answering~\cite{diefenbach2019qanswer}. 
This work builds upon Wikidata, a public and open source knowledge graph currently relying on $23,362$ active users\footnote{https://www.wikidata.org/wiki/Wikidata:Statistics} to ingest and maintain its knowledge. 
Wikidata has grown to a knowledge graph with an impressive size with  $17$ billion triples with $10,329$ properties, collecting structured information about people, places, films, stars, publications, proteins, and many more.  
Finding, modeling, monitoring, and organizing this data in a knowledge graph requires considerable work from human editors as a whole.

For a single editor, it is certainly difficult to keep track of the knowledge that is published on the Web and to follow the information that could be changing over time. The sheer volume and complexity of the data make it a demanding and time-consuming task.
For instance, \texttt{Evzen Amler} is a researcher who corresponds to the Wikidata item Q48428974\footnote{https://www.wikidata.org/wiki/Q994013}. At the same time, this researcher has the ORCID 0000-0002-0977-8922\footnote{https://orcid.org/0000-0002-0977-8922}. There might be information published on \href{https://orcid.org/}{https://orcid.org/} by the researcher that is not on Wikidata, like the affiliation or information about the education. Similarly, this could happen for Google Scholar or Facebook. It is difficult for editors to identify this external information and to monitor their modifications over time. Another example is the music group \texttt{Deskadena} that corresponds to the Wikidata item Q113585063\footnote{https://www.wikidata.org/wiki/Q113585063} which has also a link to Musicbrainz where it has artist ID f6afb1cc-8799-41cf-8fa8-2745eeab36e6\footnote{https://musicbrainz.org/artist/f6afb1cc-8799-41cf-8fa8-2745eeab36e6}. MusicBrainz might contain information like the founding year or the country that is not contained in Wikidata. Needless to say that it is difficult for editors to locate this information by human effort and insert it in Wikidata.

A common approach to alleviate these issues is the development of domain-specific scrappers. 
The first bottleneck is the need to create scrapers one by one and maintain them, which hardly scales. 
A second crucial issue is their inability to extract ``subtags expressions''. 
For instance, in Musicbrainz Web page\footnote{https://musicbrainz.org/artist/f6afb1cc-8799-41cf-8fa8-2745eeab36e6} of \texttt{Deskadena}, the founding year is provided in HTML as follows:  

\noindent \textit{<dd class="begin-date">1997<!---->(25 years ago)</dd>}

\noindent In the DOM tree, the actual content between 
\textit{<dd>} and \textit{</dd>} tags would be the text span ``\texttt{1997<!-- --> (25 years ago)}'' which is not the expected answer, that is, \texttt{1997}. This concrete problem can be solved using a regular expression provided it is hard-coded for that property. Moreover, in the broader textual value case, such extraction may yield a larger text span rather than the actual answer that we are looking to integrate into Wikidata. For example, extracting the employer of \texttt{Yann LeCun} from his Google Scholar page using Wikidata property $P1960$\footnote{\texttt{https://www.wikidata.org/wiki/Property:P1960}} (\emph{Google Scholar author ID }) and a tag-based scraper would yield the text span ``Chief AI Scientist at Facebook \& Silver Professor at the Courant Institute, New York University''. This candidate is broader than the expected answer (``Facebook" or `` New York University") and also contains his occupations (Wikidata property \texttt{occupation}\footnote{\url{https://www.wikidata.org/wiki/Property:P106}}) which is not desirable and prone to downstream process errors.

In this paper, we propose to explore this problem in a completely different manner. We cast domain-specific scraping as an open Question Answering (henceforth denoted as QA) task in order to identify new facts with high accuracy. We present a machine learning framework that uses Wikidata as a seed to train neural networks to identify new structured knowledge on external Websites. The extraction framework relies on the external identifiers provided by Wikidata. Our approach does not need any additional training data, except the data already contained in Wikidata that can serve as a seed to train our algorithms. We demonstrate that this pipeline adapts to multiple websites and can extract facts that are contained in structured, semi-structured, or unstructured parts of the external website without any particular adaptation.\\
Wikidata contains $7,424$ external identifiers which point, via the \texttt{formatter URL} property P1630, to $7,220$ unique domains. We estimate that we can extract millions of facts using this technique.

Our main contributions are the following:
\begin{itemize}
    \item We show that language models can be trained to extract facts from HTML content without additional labeling than what is already offered by Wikidata by casting the Web scraping task as an extractive Question Answering one,
    \item We show in different experiments the zero-shot, few-shot and supervised learning performance of our extraction framework.
    \item We propose an extraction framework that covers both structured, semi-structured, and unstructured Web data provided in HTML content,
    \item We publish \textit{RoBERTa-Base-WebExtractor}, a deep learning model adapted to perform Web extraction in zero and few-shot scenarios with high accuracy.
    \item We present a technique to link textual evidence to Wikidata entities.
\end{itemize}

The paper is structured as follows. In section \ref{sec:related} we describe related works. In section \ref{sec:framework} we describe our approach for question answering based on fact HTML extraction. In section \ref{sec:experiments} we present different experiments showing the performance of our approach for different domains, properties, and training data availability scenarios. In section \ref{sec:completness} we give an estimation of how many facts can be extracted via this approach. We conclude with section \ref{sec:conclusions}.

\section{Related Work}\label{sec:related}

This work is mostly related to two different areas of research namely knowledge graph completion and Web Extraction. knowledge graph completion encompasses KG completion via link prediction and KG completion via free-text. The first aims at inferring new knowledge from the KG itself by reasoning over existing knowledge~\cite{lao2011random,riedel.2013,neelakantan2015compositional,garcia2017kblrn,chen-etal-2018-variational}. For example a newspaper located in Paris has some high probability to be a newspaper in French. The second one tries to extract knowledge from text corpora like Wikipedia or the Web. This includes works like NELL~\cite{carlson2010toward} or the knowledge vault~\cite{dong2014knowledge}. The first is tailored only to text resources, while the second offers different extractors for different types of information (free-text, HTML tables, ...). Both approaches aggregate and score text snippets to extract the knowledge.

Besides knowledge graph completion and Web extraction, there exist two prior works that make use of question-answering technologies to extract knowledge from free text, namely~\cite{levy2017zero} and~\cite{kratzwald-etal-2020-intkb}. The  extraction task is applied to a text document. For example, the extraction of the property "place of birth" can be expressed as the task of answering the question "Where was [ENTITY] born?". Both works use Wikipedia as a text corpus to extract knowledge. In our work, we expand this document-level extractive QA technique to extract knowledge over HTML websites and generalize it on several Web domains.
This is, to our knowledge, the first time this approach is applied to HTML content, which is scientifically challenging for several reasons, including code diversities and lack of annotated training data.

Finally, other popular knowledge graphs like DBpedia~\cite{mendes2011dbpedia} and YAGO~\cite{Mahdisoltani2014YAGO3AK} also largely rely on extraction frameworks to build the underlying knowledge. DBpedia is tailored to the semi-structured information contained in the info-boxes of Wikipedia. These are extracted via multiple rule-based pipelines that are matching recurrent patterns appearing in the info-boxes. Similarly, YAGO extracts the info-boxes of Wikipedia and combines them with the Wikipedia categories and with information derived from WordNet. The corresponding knowledge has a manually verified accuracy of 95\%. In this work, we do not try to extract information solely from Wikipedia, but from external Web domains which are linked to the Wikidata knowledge graph. 
    
\section{\frameworkName: A framework to extract facts from websites}\label{sec:framework}

\begin{figure*}
    \centering
    \includegraphics[width=0.90\textwidth]{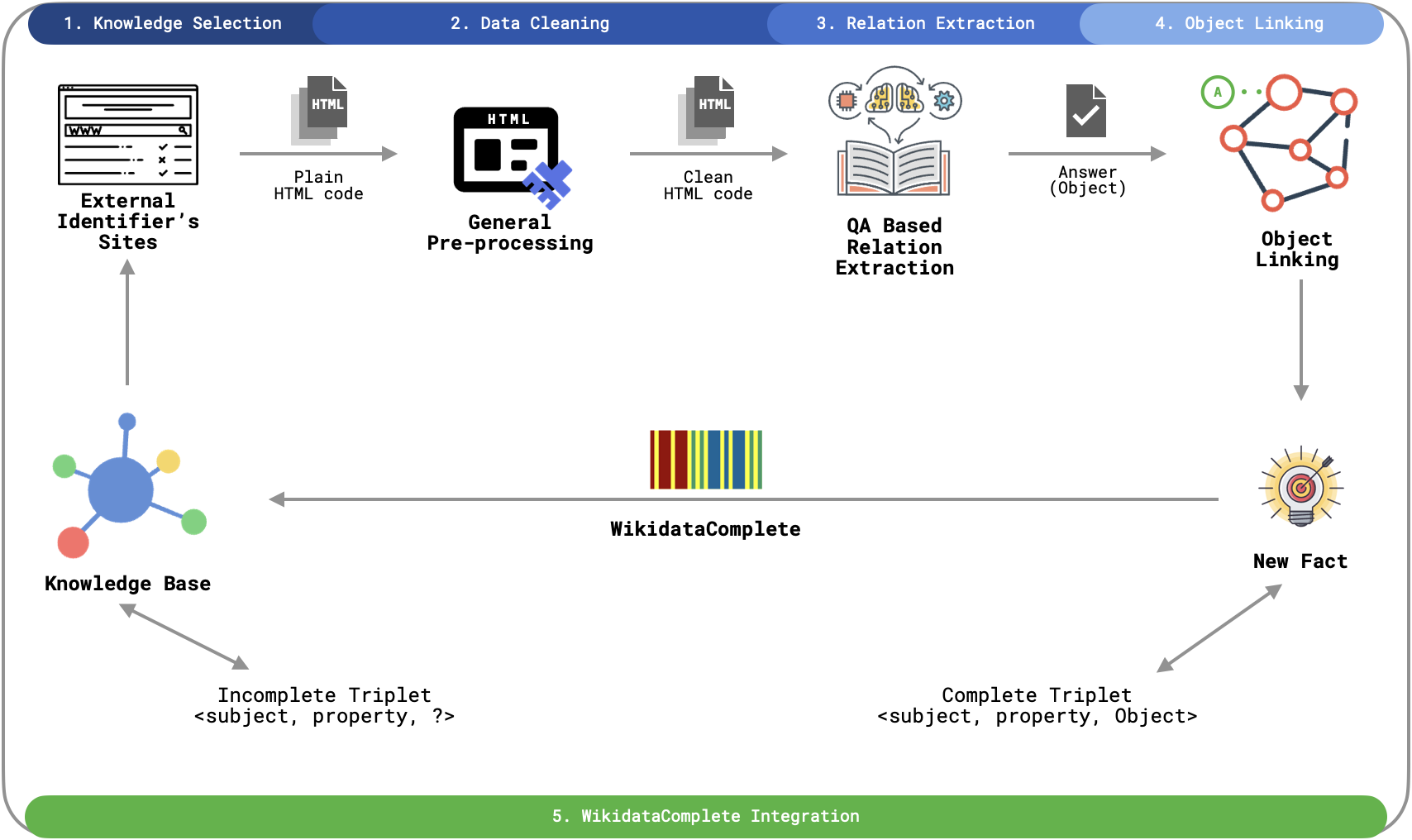}
    \caption{Diagram that demonstrates the main pipeline of our framework \frameworkName. The framework consists of different modules namely (clock-wise): knowledge selection (which identifies facts to be completed), data cleaning (which fetches websites that can contain the underlying fact and perform general cleaning),  relation extraction (which extracts the actual fact from a website), object-linking (which links the identifies object to a Wikidata item), WikidataComplete integration (which proposes extracted facts to users for fact verification).}
    \label{fig:framework}
\end{figure*}

This section describes the workflow of how we build the framework \textbf{\frameworkName} with the goal of extracting new facts from websites. The whole process is represented in Figure~\ref{fig:framework}. 

To begin with, in Wikidata, entities involved in triples can be associated with external identifiers\footnote{External identifier (different from Uniform Resource Identifiers(URI)) is a property type used in Wikidata. Properties of this type are used to point to identifiers used in external systems. Generally, these identifiers can be used to build a URL that exists on the Web and that describes the corresponding Wikidata entity.}. 
Such external identifiers point to external websites describing the corresponding Wikidata entity. 
Examples of such external identifiers include ORCID, Musicbrainz, Twitter, Facebook, and the German National Library.
Our aim is to identify subjects in Wikidata that are not complete (i.e. where some statements are missing) and identify the corresponding information in one of the associated external websites.

In the following, we describe the process in detail. We assume that we want to complete Wikidata using information from an external identifier $x$ which is pointing to an external domain. This is indicated by the presence of the property \texttt{formatter URL} (P1630) which indicates how to map the external identifier to an external URL. As a running example, we take the external identifier ORCID (P496) which has as a \texttt{formatter URL} value of "https://orcid.org/\$1".

\subsection{Knowledge Selection} \label{subsec:knowldge selection}
We first want to identify which properties can be completed using the external resources associated with $x$. We use the following rationale: entities associated with an external identifier $x$ have some common patterns. For example, entities with the ORCID external identifier are researchers and share some properties that one can probably identify on the corresponding ORCID page.

Therefore we extract a sample of entities $e_{1}, ...., e_{n}$ which have $x$ as an external identifier. In our example, the entity $e$ is "Evzen Amler" (Q994013) $x$ is ORCID. Then, for each of the identified entities we consider all properties associated with them, i.e. all properties $p$ for which there is a triple $(e_{i}, p, o)$, here $o$ represents the object in the triple. We take all these properties and list them in ascending order based on their usage between $e_{1}, ...., e_{n}$ and obtain a list of properties $\mathcal{P}$ $\langle p_1, p_2, ..., p_k \rangle$. For our running example, this list contains the property "employer" (P108) and the property "given name" (P735). 

Our aim is therefore to complete Wikidata entities $\mathcal{E}$ which have $x$ as associated external identifiers but which are missing one of the properties in $\mathcal{P}$. In our running example, we want to complete "Evzen Amler" (Q994013) with properties like "employer" (P108) and "given name" (P735).

\subsection{Data cleaning} \label{subsec: web scraping}
For the sake of completion, we report here our two-fold data cleaning approach, including curation of Web domains and HTML clean-up. 

\textbf{Curation of the Web domains:} for each of the external identifiers we want to complete, we use \textit{Selenium} library to download the full content of the websites. During the curation of websites, we skip the domains which blocked our requests. 

\textbf{HTML clean-up:} due to the diversity of the Web design patterns and libraries applied, the plain HTML content of the websites can differ a lot. Therefore we apply a general pre-processing step to the websites. This includes extracting the body node from the HTML structure, removing tags that do not contain important information (such as <script>, <style>, and <img>), and also replace the unknown tags with <start>, <end> tokens. 

\subsection{Relation extraction via QA} \label{subsec:relation extraction via qa}
We solve the Web extraction task by mapping it to an extractive question-answering task. This task identifies for a given question and context the start and end position of the answer in the context. We take inspiration from similar works~\cite{levy2017zero, kratzwald-etal-2020-intkb} over purely textual document collections like Wikipedia.

The input is an incomplete triplet $(s, p, \_)$ where $s$ represents the subject entity in the triplet and $p$ represents the property or so-called relation between subject and object. Also, the input is accompanied by an external website about the subject entity in plain HTML format. Based on these, our goal is to find the missing object entity. Therefore, to fill the blank in the triplet, we consider the problem as a QA task, where the website serves as the context and the semantic names of the property serve as questions after a simple reformulation. More concretely, we generate the questions as follows: first, we find all semantic names of the property, then we compose the question simply with each name and a question mark $?$ with a space between them. In our running example, an incomplete triple is  ("Evzen Amler", "employer", \_) and one of the corresponding questions is "employer ?".

To solve the QA task, we use a pre-trained language model and append a dense layer that is adapted to solve the QA task. Furthermore, we fine-tune it on SQuAD~\cite{rajpurkar2016squad}, a large QA dataset in order to have a baseline QA model. The experiments will show that this model is unable to extract facts from websites. The trained model is able to solve the QA task, but it is not trained to read websites, only plain text. Therefore, we extract from Wikidata additional training data to further fine-tune our model. The training data is generated as follows for a pair $(x,p)$ of an external identifier and a property. We extract from Wikipedia a sample of triples of the form $(s_{i},p,o_{i})$ where $s_{i}$ is an entity with associated external identifier $x$ pointing to a website. By string matching, we search in the website labels and aliases of $o_{i}$. If we find a mention of $o_{i}$ we use this as a training example for the QA task. Note that this process is unsupervised and uses the information that is already contained in Wikidata to construct training data adapted to the task.

The training data is used to further fine-tune the SQuAD fine-tuned language model. In the experiment section~\ref{sec:experiments} below we analyze the performance of the model in different training scenarios, for different domains and different properties.

At prediction time, we formulate the question for the incomplete triple in the same way as at training time. We use our pre-trained model to pinpoint a good answer from the websites. In the following, we want to describe 3 examples of successful extraction. 

\begin{figure}
    \centering
    \includegraphics[width=\linewidth]{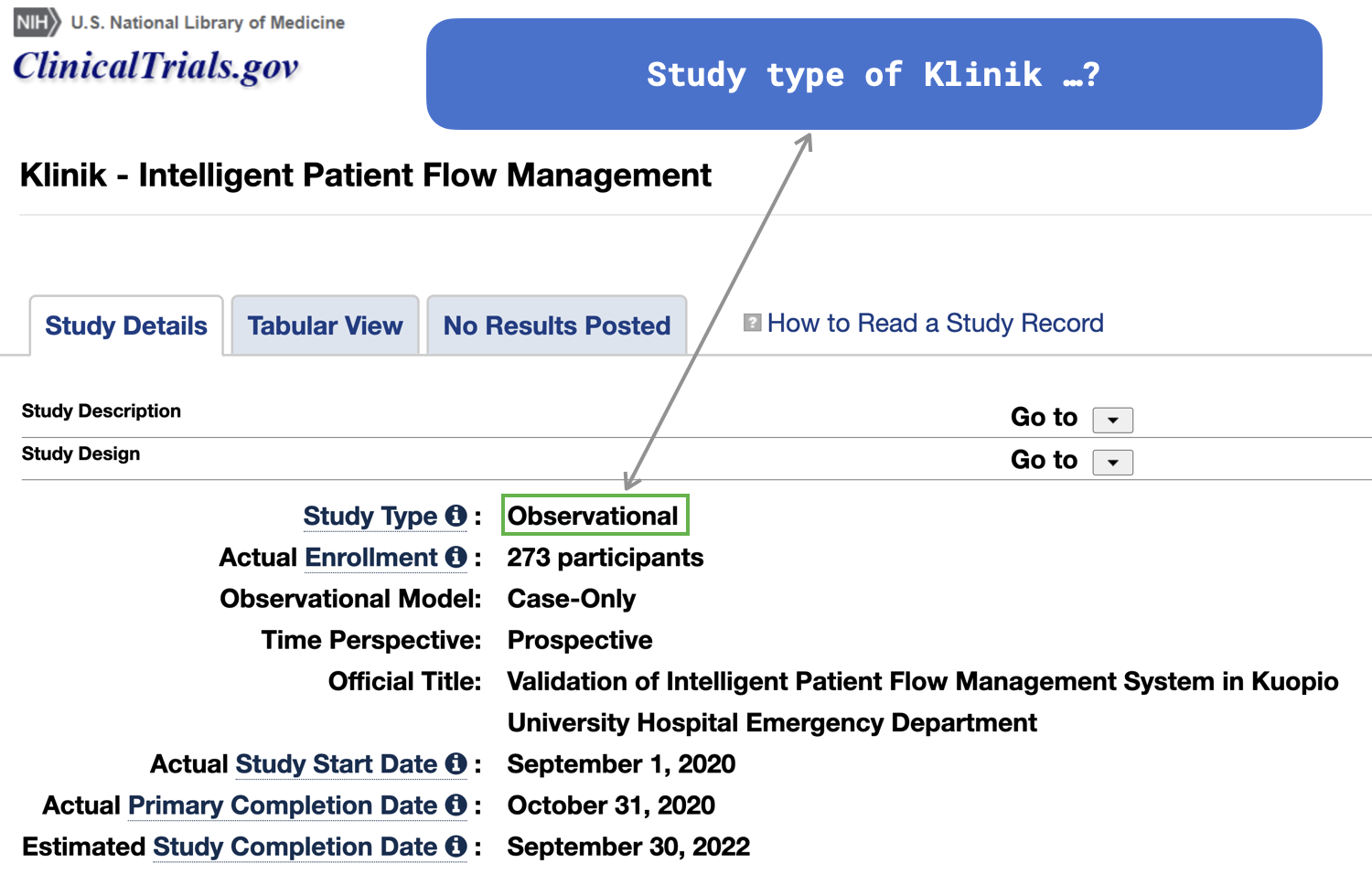}
    \caption{Web extraction from a well-structured field in the website Clinicaltrials.gov. The "study type" for the clinical trial "Klinik - Intelligent Patient Flow Management" is extracted.}
    \label{fig:structured example}
\end{figure}

\textit{Structured example:} among the websites from domain name \texttt{clinicaltrials.gov}, all have a fixed structured format storing the information. Figure~\ref{fig:structured example} shows how the property P8363 (Study type) is completed. The goal of the QA model is to find 'Observational' as the answer.
\begin{figure}[h]
    \centering
    \includegraphics[width=\linewidth]{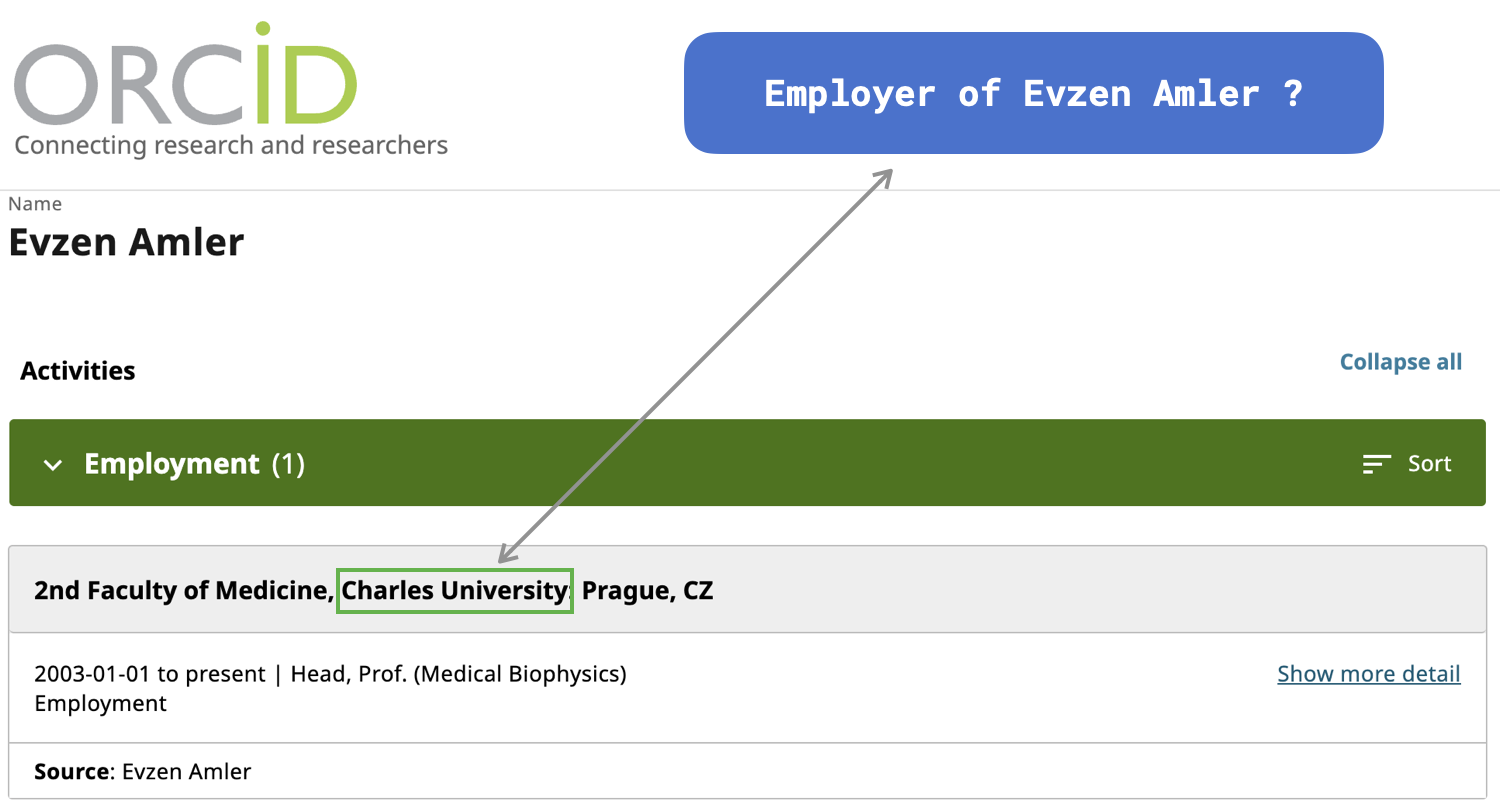}
    \caption{Web extraction from a semi-structured field in the website ORCID. We extract the "employer" of the researcher "Evzen Amler".}
    \label{fig:semi-structured example}
\end{figure}
\begin{figure}[h]
    \centering
    \includegraphics[width=\linewidth]{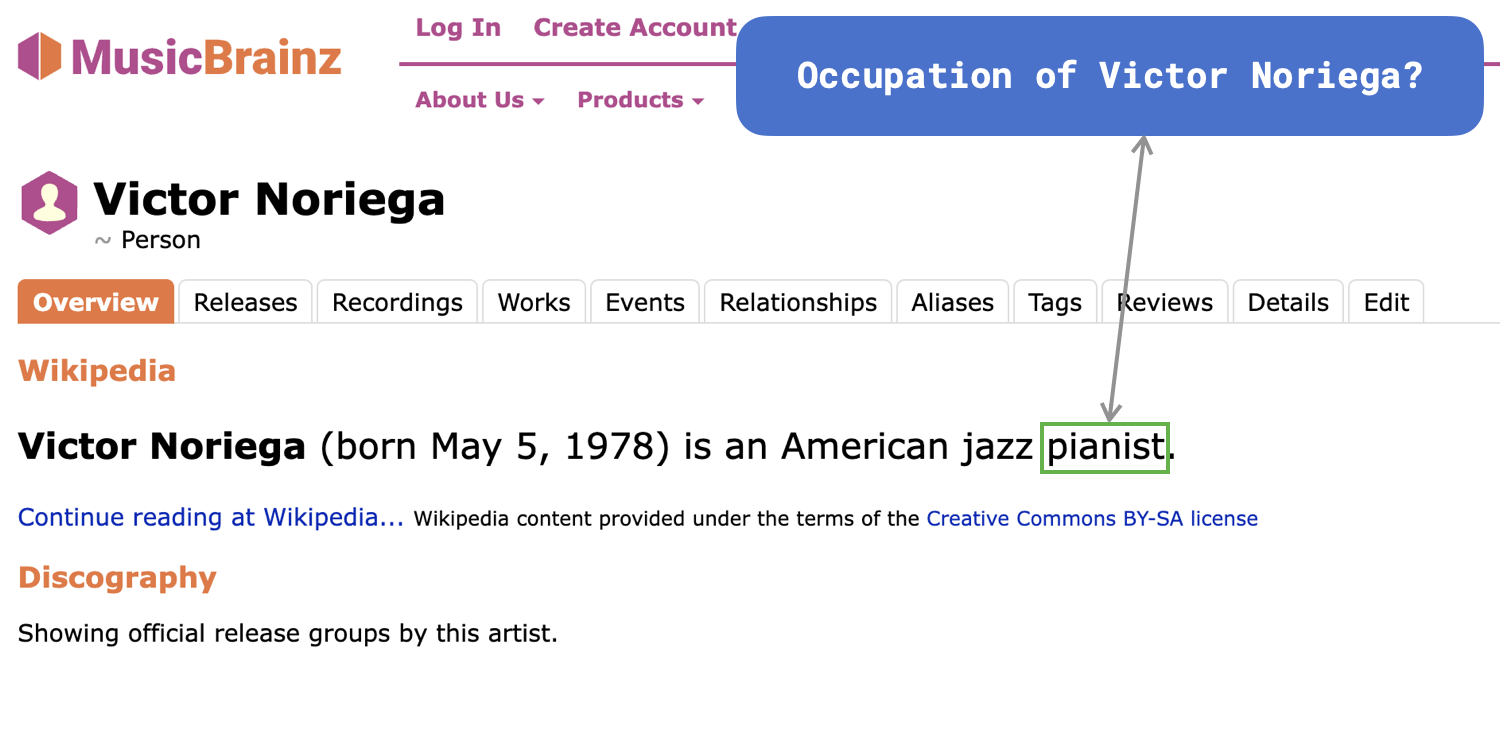}
    \caption{Web extraction from an unstructured field in the website MusicBrainz. We extract the "occupation" of "Victor Noriega".}
    \label{fig:un-structured example}
\end{figure}

\textit{Semi-structured example:} the websites from the domain \texttt{orcid.com} also have a fixed structured format for the data, but the layout of the data display makes the information semi-structured. As shown in Figure~\ref{fig:semi-structured example} the employer of "Evzen Amler" is "Charles University". However, the website presents this information as "2nd Faculty of Medicine, Charles University, Prague, CZ". Therefore, the goal of the QA model here is to locate firstly the correct position from the website, and then extract the right answer "Charles University" (and not the whole string displayed for the employer's name).

\textit{Unstructured example:} the websites accessible under the domain \texttt{musicbrainz.org} are composed of structured information as well as unstructured information. As shown in Figure~\ref{fig:un-structured example}, it is not possible to extract the property P106 (Occupation) from structured fields. Using our QA model fine-tuned for HTML extraction, we are able to find that the "occupation" of "Victor Noriega" is "pianist".

\subsection{Object Linking}
In order to ingest information in Wikidata, one needs not only to identify the textual evidence but also to link it to the correct entity in Wikidata. The biggest problem with this task is that the textual entity can be ambiguous. As a concrete example, imagine that while searching for the value of the property "educated at" (P69) we identify the textual evidence "Oxford". In Wikidata there are multiple entities that are named Oxford, like the "collegiate research university in Oxford" (Q34433) which is the one that we are searching for, the "city in Oxfordshire" (Q34217), and the "association football club in Oxford" (Q48946). In these cases, it is clear that we would like to predict the university since a person is neither educated by a city nor a football association. Based on this idea we designed the following linking strategy for textual evidence "t" and a given property $P_{i}$. We construct a machine learning-based linker with the following training strategy:
\begin{itemize}
    \item We extract from Wikidata a sample of the objects of $P_{i}$, that we denote with $n_{1}, ....,n_{k}$. For the property P69, for example, the object Q34433, i.e. Oxford the "collegiate research university in Oxford".
    \item For each object $n_{i}$ we identify all nodes in Wikidata with the same label or alternative label. We denote these nodes as $m_{i,1}, ...., m_{i,l}$. We assume that $n_{i}=m_{i,1}$ For our running example these are for example the entities Q34217 "city in Oxfordshire" and Q48946 "association football club in Oxford".
    \item For each of the nodes $m_{i}$ we collect all items that are connected to them via an outgoing link. For Oxford (Q34433) this is for example the concept of "public university" (Q875538), "Radcliffe Science Library" (Q7280038), and so on. We consider each of these items as a feature for a learning-to-rank task.
    \item We train a learning-to-rank model which ranks the features associated to $m_{i,1}, ...., m_{i,l}$ so that $m_{i,1}=n_{i}$ is ranked first. The idea is that the learning to the rank model should learn that a "public university" has a higher chance to be a good candidate than a "football club". This is exactly following the intuition we presented above.
\end{itemize}
Once the property is trained, we simply follow the same procedure used at training time to predict for textual evidence "t" the entity in Wikidata that is most likely corresponding to it.

\subsection{WikidataComplete Integration}
\begin{figure}[h]
    \centering
    \includegraphics[width=1.0\linewidth]{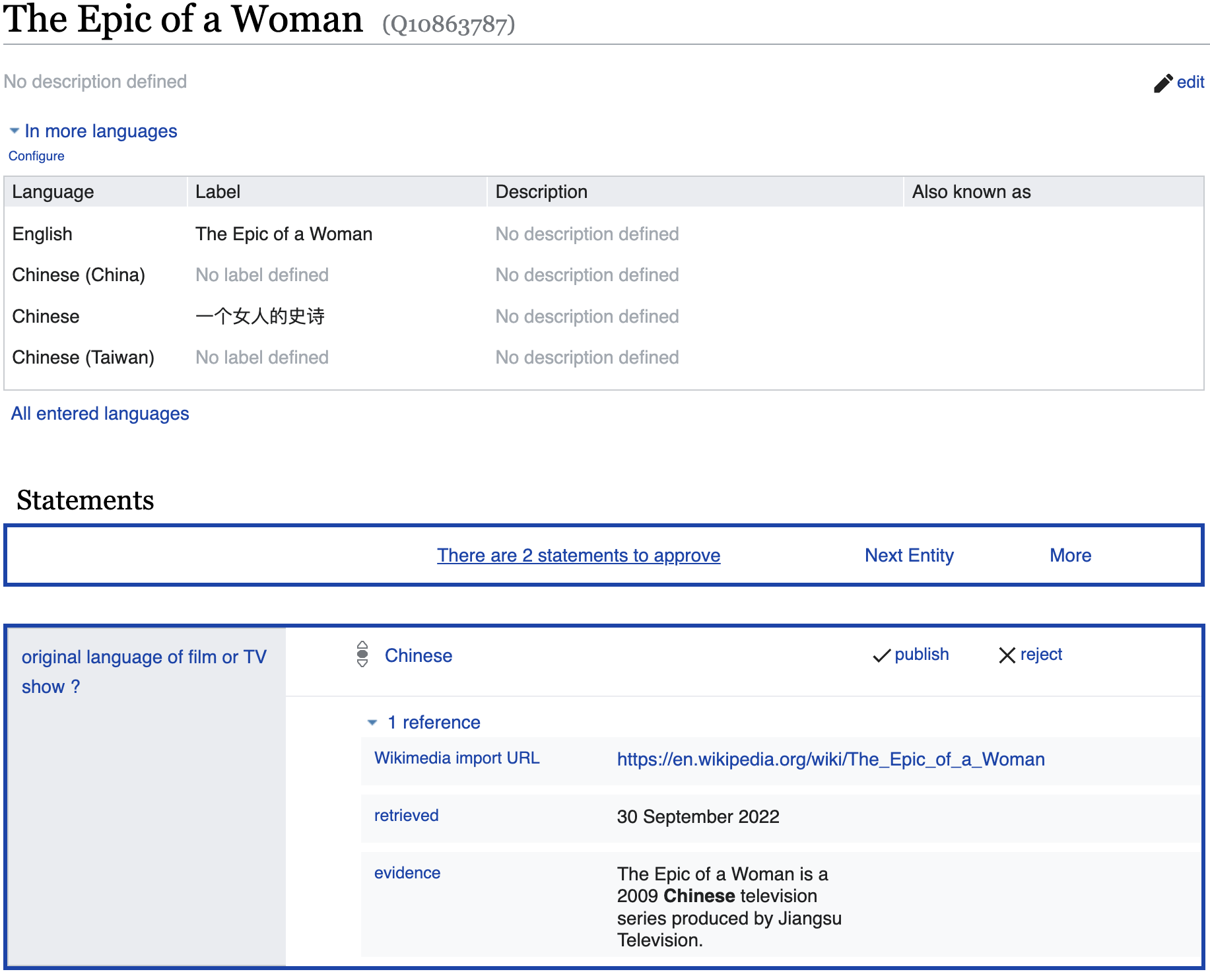}
    \caption{WikidataComplete: a Wikidata gadget that is intended to help users in adding more facts to the Wikidata knowledge base. In the statement section, a user can see statements to approve or reject. A reference is given in order to understand on which website the fact was found and what is the evidence for the underlying fact.}
    \label{fig:wikidatacomplete gadget}
\end{figure}

The extracted facts might be of different quality. Depending on the extraction domain and the property, the extraction can be simpler or more difficult. Errors are inevitable in both cases. To avoid introducing errors to Wikidata, we propose not to ingest the predicted facts directly to Wikidata, but to validate them by Wikidata editors. Therefore, we donate the facts to WikidataComplete~\cite{10.1007/978-3-031-11609-4_22}, an easy-to-use and user-friendly Wikibase plugin that presents new facts to Wikidata editors. A fact is presented directly on the Wikidata page corresponding to the subject of the fact. A screenshot is shown in Figure~\ref{fig:wikidatacomplete gadget}.

\section{Experiments}\label{sec:experiments}
In this section, we perform an exhaustive experimental analysis of our framework \textbf{\frameworkName}.

Given a domain $D$ and a property $P$, that we want to complete with the information contained in $D$, different situations can occur.
We consider the following:
\begin{itemize}
    \item \textbf{Supervised Learning}: this situation occurs if we can generate from $256$ to $500$ examples for a domain/property pair ($D$, $P$). Note that this is a relatively frequent case. If Wikidata has good completeness with respect to property $P$ we can easily collect this amount of examples in the unsupervised way we described in Section \ref{sec:framework}.
    \item \textbf{Few-Shot}: this scenarios occurs if we can generate from $8$ to $128$ examples per domain and property. This occurs in situations when the property $P$ is very incomplete, or when entities linked to the domain are incomplete with respect to $P$. 
    \item \textbf{Zero-Shot}: for new external identifiers, it is possible that one can not collect any existing training data from Wikidata. This can happen if the property was newly created or if all entities linked to the domain $D$ do not contain the property $P$.
\end{itemize}

To take into account all of these situations we designed different experiments that study the performance of the approach we propose. We fine-tuned in total $512$ QA models and conducted $1,737$ evaluation runs to study how our framework performs in different experimental settings. \\
\begin{table}[h]
    \centering
    \resizebox{1.0\linewidth}{!}{
        \begin{tabular}{ccccc}
        \toprule
             \multirow{2}{*}{\textbf{\# of Domains}} & 
             \multicolumn{2}{c}{\textbf{\# of properties}} & 
             \multirow{2}{*}{\textbf{\# of train data}} &
             \multirow{2}{*}{\textbf{\# of test data}} \\
             & Avg & Total & \\
        \midrule
             54 & 2.981 & 161 & 500  & 500 \\
        \bottomrule
        \end{tabular}
    }
    \caption{Statistics of the dataset we used in the experiments. We indicate the number of domains, the number of properties, and the amount of training and test data}
    \label{tab:dataset_stats}
\end{table}

\subsection{Statistics of the dataset} \label{subsec:exps-statistics}
We collected our training dataset using the methods described in Section~\ref{subsec:knowldge selection}. From the top 200 most frequent external identifiers, we select $54$ unique Web domains. We excluded many domains since  website scraping is often challenging due to technical hurdles put in place by the website providers -- this includes Linkedin or Facebook for example. \\
For each domain $D$ we computed the most used properties between all entities that are linked to the domain $D$. For example for all entities in Wikidata linked to MusicBrainz, frequent properties are "occupation" ($P106$), "place of birth" ($P19$), "given name" ($P735$). To be able to investigate all 3 scenarios (supervised, few-shot, zero-shot) we restrict to properties for which we would generate at least $1,000$ examples, $500$ for training, and $500$ for testing. In the case of the MusicBrainz domain these are the properties "sex and gender" ($P21$), "occupation" ($P106$), "country of citizenship" ($P27$),  "genre" ($P136$) for which we have $1,000$ examples each, $500$ for training and $500$ for testing.\\

On average, our dataset contains $2.981$ properties for each of the considered domains. Statistics about the dataset we created are shown in Table~\ref{tab:dataset_stats}. The datasets will be published online for downloading after the review period.
%
%
%

\begin{figure*}
    \centering
    \includegraphics[width=1.0\textwidth]{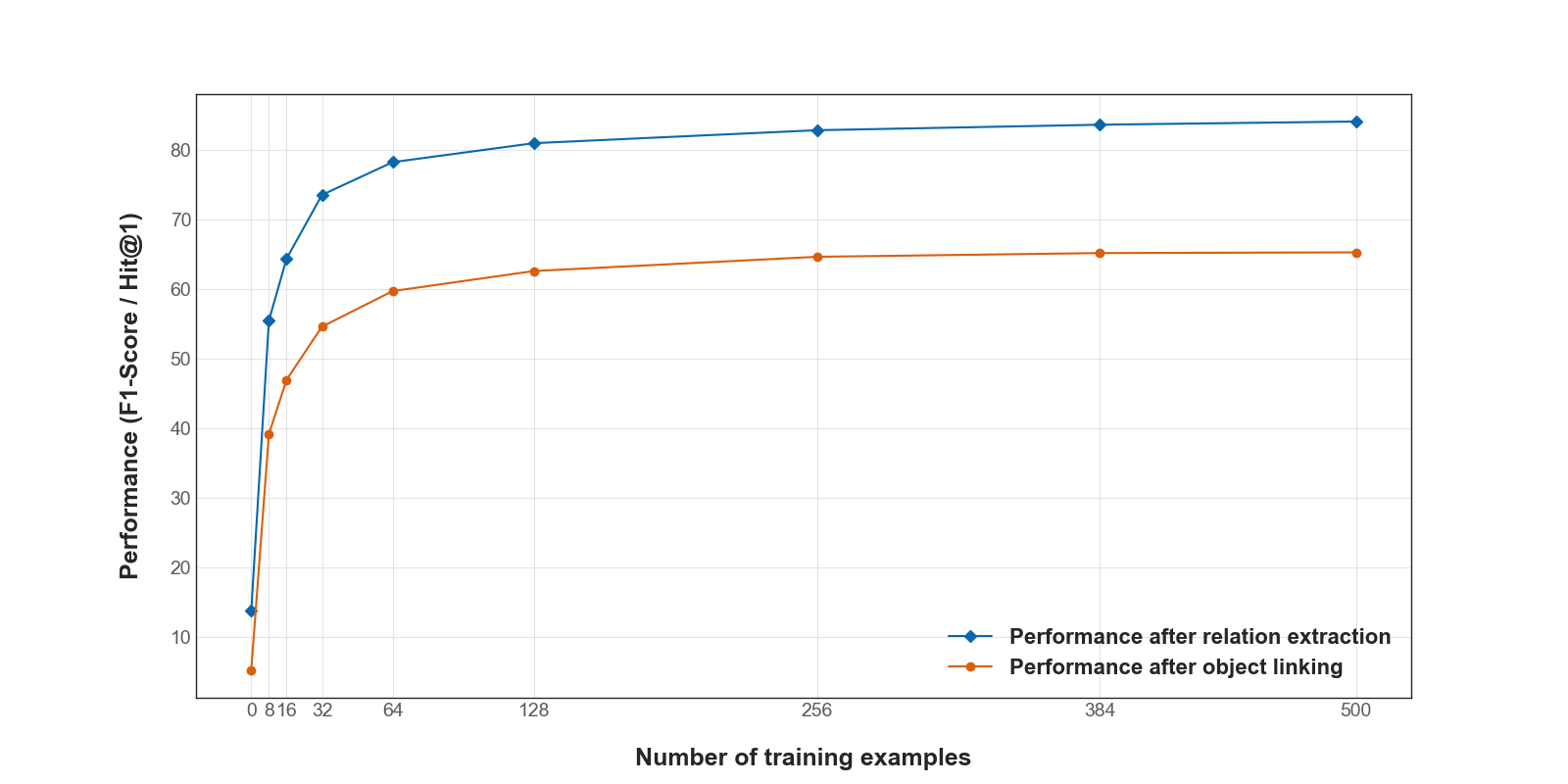}
    \caption{Performance of the framework after the relation-extraction (blue curve) and the object-linking step (orange curve) for different amounts of training data when evaluated on  54 domains. We show the performance for zero-shot ($0$ examples), few-shot ($8$, $16$, $32$, $64$, $128$) and full-training ($256$, $384$, $500$) scenarios.}
    \label{fig:annotation budget from 0 to 500}
\end{figure*}
\subsection{Experiment Settings} \label{subsec: experiment-settings}

For the baseline QA model, we choose the pre-trained, 12-layers,
768-hidden, 12-heads, 125M parameters language model \textit{RoBERTa-Base}~\cite{liu2019roberta} fine-tune on the SQuAD~\cite{rajpurkar2016squad} dataset. 
Before further fine-tuning, we append to the vocabulary of the language model some tokens to represent HTML basic tags such as \texttt{<div>}, \texttt{<p>}, \texttt{<h1>} to \texttt{<h6>}, \texttt{<ul>}, \texttt{<li>}. Besides, we also add \texttt{<start>} token and  \texttt{<end>} token to represent all the boundaries of unknown tags.

We use the \textit{AdamW} optimizer with a learning rate $ 2e-5 $ and batch size $ 48 $. All QA models were fine-tuned for $ 2 $ epochs.

As evaluation metrics, we report \textit{F1-score} for the performance of the QA models and \textit{Hit@1} for the object-linking performance which is also the final end-to-end performance.\\

\subsection{Experiment Results}
%
In what follows, we present and discuss the results obtained in our experiments for the different scenarios mentioned above.\\

\textit{Supervised Learning}
\label{exp-supervised-learning}
In this setting, we take the \textit{RoBERTa-Base} language model pre-trained on SQuAD and we fine-tune it on our dataset prepared on $54$ external domains. We fine-tune a QA model separately for each of the external domains but by merging the training data for all properties (note that for each domain we have on average $2.981$ properties). The goal of this fine-tuning step is that the model learns to extract information from a specific domain and at the same time learns to extract specific properties. Preliminary experiments show that training for all properties at the same time is beneficial because the language model has more training data to adapt to a specific domain.

After fine-tuning the QA model in our framework with the created dataset, we achieve across the $ 54 $ domains an average F1-score of $ 84.07 $ on the test set.
This high performance demonstrates that our framework \textbf{\frameworkName} can extract from a wide range of different domains and properties.

We report the results for each domain/property pair in Table~\ref{tab:full_finetune_part_1} and Table~\ref{tab:full_finetune_part_2} of the appendix. We report that the performance is significantly fluctuating  between domain/property pairs. 
For example, the model is not able to learn to extract the property "cast member" (P161) from the domain IMDb ID (P345) where the performance is 9.5 at F1-score. 
This is because, for movie entries in the domain IMDb, the cast members are usually more than one actor/actress. However, after fine-tuning the QA model, it only predicts the first one of the cast member list.
On the other hand, for the property sex or gender (P21) of the domain MusicBrainz (P434) the performance is 99.2 at F1-score. 
We noticed that the model is good when the pattern to learn is simple and has less good performance when the data is more unstructured.

The end-to-end performance after linking drops to $ 65.3 $ \% Hit@1. 
This is due to the ambiguity of the linked terms.
For example, under the domain Canmore ID (P718), when we search for the property value of "is located in" (P131) we find the textual value "highland". 
It actually matches the label of the object we want to fill. 
However, after the object linking step, we have the object Highlands "borough of New Jersey, United States"(Q1086265) as the linked entity, which is not the target entity Highlands "council area in the Scottish Highlands" (Q208279) which we are searching for.
%
Our analysis shows that ambiguity is the main reason that causes the drop in overall performance. \\


%

%

%

%

\textit{Zero-Shot/Few-shot Learning} \label{exp-zero-few-shot-learning}
In the following, we analyze the performance of the framework in the cases where there are no, to very few training examples. The results are presented in Figure~\ref{fig:annotation budget from 0 to 500}. Each data-point corresponds to the average performance on the $54$ domains for $0$, $8$, $16$, $32$, $64$, $128$, $256$, $384$, $500$ training examples per property -- the test set remains the same ($500$ properties). The blue line indicates the performance after the relation extraction step (F1-score), and the orange line to the performance after the object-linking step (Hit@1).

In a zero-shot scenario, the framework performs tediously. This shows that the task of Web extraction is far from the QA task on which the model is originally fine-tuned.

On the other side, the language model is able to quickly learn the new task with just a few examples. With $8$ examples, the query performance of the relation extraction (blue line) jumps from $12.23$ to $48.66$ (an increase of 36 at the F1-score). This indicates that a small annotation effort can lead to an important performance boost. There is in general a clear trend when increasing the budget size from 8 to 64. With $64$ examples the performance reaches $78$ at F1-score.
%
Taken together, these results suggest that the Web extraction task is different from traditional QA tasks, but the adaptation from the QA task to the website extraction task can be achieved via few-shot learning.
After adding more than $ 128 $ training examples per property, there is no significant improvement in the performance.
The performance only increased by $ 2.65 $ \% with a $\times 3$ annotation effort in training data collection. \\

\textit{Pre-trained QA model for Web extraction.} \label{exp-pre-trained-model}
Overall the language model needs to learn:
\begin{enumerate}
    \item The task of Web extraction is far from the QA task on which it is fine-tuned as shown in our experiments.
    \item Patterns that are specific to a domain
    \item Patterns that are specific to the property that is extracted.
\end{enumerate}

While (2) and (3) are knowledge that can only be injected into the language model with suitable training data, one can try to create a language model that is more suited to (1). Our experiments in the zero-shot and few-shot scenarios also show that a language model is able to learn quickly the task.
To assess how pre-training the language model for Web extraction can affect our framework, we design an additional experiment. We split our dataset consisting of $ 54 $ domains into 2 parts. We use 44 domains (and all their properties) to train a language model for the Web extraction task and left 10 domains (and all their properties) for testing. This corresponds to an 80/20 split of the overall dataset. The objective is to estimate the performance of a model that would be trained on (close to) to all available Wikidata information for the community to use on new unseen domains. \\
%


The experimental setup is as follows: 
\begin{itemize}
\item First we pre-train the baseline QA model with all training data we collected for the $44$ domains which we denote as \textit{RoBERTa-Base-WebExtractor}. 
\item  Second we repeat the experiments we did in the few-shot learning setup using \textit{RoBERTa-Base-WebExtractor}. The evaluation will be performed on the hold-out set only. 
\item Third, we compare the results we collected before in the few-shot learning and supervised learning experiments using \textit{RoBERTa-Base} fine-tuned on SQuAD (by restricting only to the $10$ domains we kept for testing).
\end{itemize}

The result is shown in Table~\ref{tab:with_vs_without}. The left part contains the results for the baseline QA model while the right part contains the results for the language model pre-trained for Web extraction.

From Table~\ref{tab:with_vs_without}, we observed a significant performance boost after pre-training the QA model. 
As discussed above, the QA models fail to perform the Web extraction task in the zero-shot scenario.
However, with pre-training, the \textit{RoBERTa-Base-WebExtractor} model is able to achieve a $38.08$ F1-score boost in the zero-shot setting.
It is worth mentioning that with only $8$ examples of few-shot fine-tuning, \textit{RoBERTa-Base-WebExtractor} is able to achieve $64.91$ F1-score against $79.85$ F1-score in the case of full training with 500 examples. This means that the model is able to achieve with only $8$ examples already $ 82.29\% $ of the performance the fully trained model can achieve.

%
The results in this section indicate that, with \textit{RoBERTa-Base-WebExtractor} as the starting point, the framework can achieve good performance already with very few examples. 
This is, in particular, beneficial when one needs to perform Web extraction in zero-shot and few-shot scenarios.
We will publish the pre-trained model \textit{Roberta-Base-WebExtractor} for downloading and further research usage after the review period.



%

\begin{table}[H]
    \centering
    \resizebox{1.0\linewidth}{!}{
        \begin{tabular}{ccc|cc}
        \toprule
         & 
         \multicolumn{2}{c}{\textbf{Without Pre-training}} & 
         \multicolumn{2}{c}{\textbf{With Pre-training}} \\
        \cmidrule{2-5}
         Budget (K=)   &
         After RE  &
         After OL  & 
         After RE  &
         After OL  \\
         \midrule
            0  & 12.23 & 0.05 & 50.31 & 36.24 \\
         \midrule
            8  & 48.66 & 35.67 & 64.91 & 49.44  \\
           16  & 55.67 & 40.84 & 70.68 & 55.99  \\
           32  & 65.34 & 50.18 & 73.26 & 58.55  \\
           64  & 71.31 & 55.67 & 74.81 & 60.02  \\
           128 & 73.61 & 58.14 & 77.00 & 61.76  \\
         \midrule
          256  & 78.37 & 63.04 & 78.95 & 63.55  \\
          384  & 78.78 & 63.57 & 79.53 & 64.68  \\
          500  & 79.54 & 63.87 & 79.85 & 64.53  \\
         \bottomrule
        \end{tabular}
    }
    \caption{Evaluation on the hold-out set (10 domains): we evaluate the performance of QA models with and without pre-training on 10 new domains QA models haven't seen before. We present the performance from two stages inside the pipeline: after relation extraction (After RE) and after object linking (After OL).}
    \label{tab:with_vs_without}
\end{table}

\section{Estimation of potential contribution for Wikidata}\label{sec:completness}
In this section, we estimate the number of facts that can be contributed via \textbf{\frameworkName} to Wikidata.
We consider the $54$ domains from the dataset we created -- since we remove domains that can be hardly crawled, the estimate will be a lower bound. For each domain/property pair $(D,P)$, we can calculate how many entities in Wikidata linked to a website in $D$ have property $P$. We denote this by $\#links_{(D,P)}$ For example there are $223,281$ Wikidata entities linked to MusicBrainz (external identifier $P434$). Of these $65,074$ have no associated property "sex or gender" ($P21$), i.e. are incomplete with respect to this property.
Moreover, for each pair $(D,P)$ we can calculate how often we find the value of $P$ in the domain $D$, we denote this ratio by $freq_{(D,P)}$. For example, if we sample 100 MusicBrainz websites we can find in $94$ of them the value of the "sex or gender".
Our experiments give the Web extraction accuracy (after object linking) for each of the pair $(D,P)$ which we denote by $acc_{(D,P)}$. After object linking the accuracy for $(Musicbrainz, P21)$ is $ 19.4\% $.
We can therefore estimate the total number of facts we can complete using the following formula:
$$
\#facts_{(D,P)} = \#links_{(D,P)} \cdot freq_{(D,P)} \cdot acc_{(D,P)}
$$
For $(Musicbrainz, P21)$ we can expect to extract $65,074 \times \frac{94}{100} \times 19.4\% = 11,866$ facts. Overall our estimations show that just for the $54$ considered domains we should be able to extract $7,543,444$ facts. Considering that Wikidata has $7,220$ external identifiers, we estimate that we can extract millions of new facts from the Web for approval to Wikidata editors.

\section{Conclusion}\label{sec:conclusions}


This study set out to simplify the process of finding, monitoring and organizing heterogeneous data from the Web to build or complete highly structured data repositories.

For this matter, we presented \textbf{\frameworkName}, a framework that uses Wikidata as a seed to extract facts from the Web. We have shown that by using QA technologies it is possible to generate, with distant supervision, extractors for website domains connected to Wikidata to aggregate knowledge across the Web. Despite hand-crafted extractors, QA extractors:
\begin{enumerate}
\item Can be trained using the data already present in Wikidata and does not need any human intervention,
\item Can extract not just content contained between HTML tags, but exploit the natural language understanding capabilities of language models to extract more fine-grained information.
\end{enumerate}

To the best of our knowledge, this is the first study to perform Web extraction using Wikidata as a seed for extractive QA, that is machine comprehension over HTML pages without human-annotated dataset but using existing Wikidata entries as seeds.

Our experiments show how this technique performs under different training data scenarios. In full-trained settings, we can achieve high performances. Moreover, our model \textit{Roberta-Base-WebExtractor} fine-tuned for Web extraction can achieve good performances also for few-shot and zero-shot scenarios.

Coupled with a new entity linking strategy we can create extraction pipelines that can discover millions of new statements in order to help Wikidata editors and make Wikidata more complete. 
We plan to extend the number of domains for extraction and then donate these to Wikidata. Moreover, we want to investigate how these techniques perform in multilingual settings and study how to close the gap between performances before and after object linkage.


\bibliographystyle{ACM-Reference-Format}
\bibliography{WWW2023}

\appendix
\section{Appendix}

\begin{table}[H]
    \centering
    \resizebox{1.0\linewidth}{!}{
        \begin{tabular}{lcccccccc}
        \toprule
            \multirow{3}{*}{\textbf{External Identifier (PID)}} &
            \multicolumn{7}{c}{\multirow{3}{*}{\textbf{F1-Score}}} &
            \multirow{3}{*}{\textbf{Domain Avg F1}} \\
            & & \\
            & & \\
        \midrule
             \multirow{2}{*}{MusicBrainz artist ID (P434)} & P21 & P106 & P27 & P136 & - & - & - & \multirow{2}{*}{78.5} \\
             & 99.2 & 55.1 & 96.0 & 63.7 & -  & -  & -  &  \\
        \midrule
             \multirow{2}{*}{MusicBrainz release group ID (P436)} & P136 & P175 & P264 & - & - & - & - & \multirow{2}{*}{78.7} \\
             & 61.0 & 92.6 & 82.6 & - & -  & -  & -  &  \\
        \midrule
             \multirow{2}{*}{ClinicalTrials.gov ID (P3098)} & P6153 & P4844 & P8005 & P8363 & P1050 & P921 & P17 & \multirow{2}{*}{57.1} \\
             & 59.5 & 44.5 & 57.3 & 88.7 & 59.4  & 24.5  & 65.9 &  \\
        \midrule
             \multirow{2}{*}{UniProt protein ID (P352)} & P682 & P703 & P702 & P681 & P680 & - & - & \multirow{2}{*}{49.5} \\
             & 15.1 & 99.9 & 85.3 & 14.6 & 32.7  & -  & -  &  \\
        \midrule
             \multirow{2}{*}{Geni.com profile ID (P2600)} & P21 & P25 & P735 & P22 & P40 & P734 & - & \multirow{2}{*}{83.4} \\
             & 89.0 & 89.0 & 94.3 & 93.2 & 45.7 & 88.9  & -  &  \\
        \midrule
             \multirow{2}{*}{Library of Congress authority ID (P244)} & P734 & P735 & - & - & - & - & - & \multirow{2}{*}{91.1} \\
             & 93.7 & 88.5 & - & - & -  & -  & -  &  \\
        \midrule
             \multirow{2}{*}{Open Library ID (P648)} & P734 & P735 & P1412 & - & - & - & - & \multirow{2}{*}{83.8} \\
             & 91.9 & 90.5 & 67.5 & - & -  & -  & -  &  \\
        \midrule
             \multirow{2}{*}{Entrez Gene ID (P351)} & P703 & P279 & P688 & P684 & - & - & - & \multirow{2}{*}{82.4} \\
             & 98.6 & 49.9 & 92.9 & 88.2 & -  & -  & -  &  \\
        \midrule
             \multirow{2}{*}{IMDb ID (P345)} & P19 & P735 & P495 & P161 & P27 & P364 & - & \multirow{2}{*}{77.4} \\
             & 93.7 & 96.8 & 86.9 & 9.5 & 89.2 & 88.2  & -  &  \\
        \midrule
             \multirow{2}{*}{OpenStreetMap relation ID (P402)} & P131 & P150 & - & - & - & - & - & \multirow{2}{*}{60.8} \\
             & 97.8 & 23.8 & - & - & -  & -  & -  &  \\
        \midrule
             \multirow{2}{*}{CERL Thesaurus ID (P1871)} & P21 & P735 & P19 & P734 & P20 & P27 & P1412 & \multirow{2}{*}{88.0} \\
             & 95.8 & 74.8 & 88.6 & 91.6 & 86.2  & 93.2 & 85.8 & \\
        \midrule
             \multirow{2}{*}{GND ID (P227)} & P735 & P19 & P734 & - & - & - & - & \multirow{2}{*}{90.0} \\
             & 87.1 & 91.5 & 91.5 & - & -  & -  & -  &  \\
        \midrule
             \multirow{2}{*}{Tropicos ID (P960)} & P105 & P171 & - & - & - & - & - & \multirow{2}{*}{98.9} \\
             & 99.4 & 98.4 & - & - & -  & -  & -  &  \\
        \midrule
             \multirow{2}{*}{ORCID (P496)} & P108 & P735 & - & - & - & - & - & \multirow{2}{*}{85.8} \\
             & 74.5 & 97.0 & - & - & -  & -  & -  &  \\
        \midrule
             \multirow{2}{*}{EUNIS ID for species (P6177)} & P105 & P171 & - & - & - & - & - & \multirow{2}{*}{98.1} \\
             & 96.4 & 99.9 & - & - & -  & -  & -  &  \\
        \midrule
             \multirow{2}{*}{Encyclopedia of Life ID (P830)} & P105 & P171 & - & - & - & - & - & \multirow{2}{*}{96.1} \\
             & 97.6 & 94.6 & - & - & -  & -  & -  &  \\
        \midrule
             \multirow{2}{*}{Internet Archive ID (P724)} & P1433 & P407 & - & - & - & - & - & \multirow{2}{*}{99.6} \\
             & 100.0 & 99.2 & - & - & -  & -  & -  &  \\
        \midrule
             \multirow{2}{*}{IdRef ID (P269)} & P734 & P735 & - & - & - & - & - & \multirow{2}{*}{92.2} \\
             & 92.1 & 92.2 & - & - & -  & -  & -  &  \\
        \midrule
             \multirow{2}{*}{iNaturalist taxon ID (P3151)} & P105 & P171 & - & - & - & - & - & \multirow{2}{*}{94.7} \\
             & 99.0 & 90.4 & - & - & -  & -  & -  &  \\
        \midrule
             \multirow{2}{*}{PlantList-ID (P1070)} & P105 & P171 & - & - & - & - & - & \multirow{2}{*}{97.8} \\
             & 100.0 & 95.6 & - & - & -  & -  & -  &  \\
        \midrule
             \multirow{2}{*}{ADS bibcode (P819)} & P921 & P1433 & - & - & - & - & - & \multirow{2}{*}{76.6} \\
             & 54.2 & 98.9 & - & - & -  & -  & -  &  \\
        \midrule
             \multirow{2}{*}{Observation.orgID (P6105)} & P171 & P105 & - & - & - & - & - & \multirow{2}{*}{97.6} \\
             & 99.6 & 95.6 & - & - & -  & -  & -  &  \\
        \midrule
             \multirow{2}{*}{Palissy ID (P481)} & P131 & - & - & - & - & - & - & \multirow{2}{*}{100.0} \\
             & 100.0 & - & - & - & -  & -  & -  &  \\
        \midrule
             \multirow{2}{*}{ČSFDfilm ID (P2529)} & P344 & P161 & P58 & P57 & - & - & - & \multirow{2}{*}{61.9} \\
             & 88.9 & 8.6 & 65.8 & 84.4 & -  & -  & -  &  \\
        \midrule
             \multirow{2}{*}{MRAuthor ID (P4955)} & P735 & - & - & - & - & - & - & \multirow{2}{*}{98.2} \\
             & 98.2 & - & - & - & -  & -  & -  &  \\
        \midrule
             \multirow{2}{*}{Fatcat ID (P8608)} & P921 & P1433 & P50 & - & - & - & - & \multirow{2}{*}{58.9} \\
             & 47.9 & 99.9 & 29.0 & - & -  & -  & -  &  \\
        \midrule
             \multirow{2}{*}{BIBSYS ID (P1015)} & P734 & P735 & - & - & - & - & - & \multirow{2}{*}{94.6} \\
             & 98.1 & 91.2 & - & - & -  & -  & -  &  \\
        \bottomrule

        \end{tabular}
    }
    
    \caption{Experiment results of supervised learning scenario i.e. 500 examples per property per domain. (PART 1)}
    \label{tab:full_finetune_part_1}
\end{table}


\newpage

\begin{table}[H]
    \centering
    \resizebox{1.0\linewidth}{!}{
        \begin{tabular}{lcccccccc}
        \toprule
            \multirow{3}{*}{\textbf{External Identifier (PID)}} &
            \multicolumn{7}{c}{\multirow{3}{*}{\textbf{F1-Score}}} &
            \multirow{3}{*}{\textbf{Domain Avg F1}} \\
            & & \\
            & & \\
        \midrule
             \multirow{2}{*}{Olympedia people ID (P8286)} & P734 & P641 & P27 & P735 & P21 & P1344 & P19 & \multirow{2}{*}{93.6} \\
             & 95.7 & 96.2 & 96.4 & 84.9 & 99.8  & 89.1  & 92.8  &  \\
        \midrule
             \multirow{2}{*}{Ensembl gene ID (P594)} & P688 & P684 & P703 & P2548 & - & - & - & \multirow{2}{*}{97.7} \\
             & 99.4 & 91.4 & 100.0 & 100.0 & -  & -  & -  &  \\
        \midrule
             \multirow{2}{*}{SNACARK ID (P3430)} & P735 & P1412 & P734 & - & - & - & - & \multirow{2}{*}{83.0} \\
             & 83.6 & 74.2 & 91.2 & - & -  & -  & -  &  \\
        \midrule
             \multirow{2}{*}{Terezín Memorial Database ID (P9300)} & P551 & P20 & P735 & P734 & - & - & - & \multirow{2}{*}{93.8} \\
             & 94.6 & 99.4 & 99.5 & 81.5 & -  & -  & -  &  \\
        \midrule
             \multirow{2}{*}{MycoBank taxon name ID (P962)} & P105 & P171 & - & - & - & - & - & \multirow{2}{*}{98.9} \\
             & 98.0 & 99.8 & - & - & -  & -  & -  &  \\
        \midrule
             \multirow{2}{*}{OFDb film ID (P3289)} & P3578 & P1343 & P5166 & P9072 & P3432 & - & - & \multirow{2}{*}{78.5} \\
             & 30.2 & 85.5 & 90.4 & 96.9 & 89.7 & -  & -  &  \\
        \midrule
             \multirow{2}{*}{Ensembl transcript ID (P704)} & P688 & P2548 & P684 & - & - & - & - & \multirow{2}{*}{97.7} \\
             & 99.8 & 100.0 & 93.2 & - & -  & -  & -  &  \\
        \midrule
             \multirow{2}{*}{Index Fungorum ID (P1391)} & P105 & P171 & - & - & - & - & - & \multirow{2}{*}{65.9} \\
             & 32.4 & 99.4 & - & - & -  & -  & -  &  \\
        \midrule
             \multirow{2}{*}{National Diet Library ID (P349)} & P734 & P735 & P27 & - & - & - & - & \multirow{2}{*}{96.6} \\
             & 95.4 & 94.3 & 100.0 & - & -  & -  & -  &  \\
        \midrule
             \multirow{2}{*}{Libris-URI (P5587)} & P291 & P735 & P50 & - & - & - & - & \multirow{2}{*}{94.1} \\
             & 97.0 & 87.9 & 97.4 & - & -  & -  & -  &  \\
        \midrule
             \multirow{2}{*}{Discogs master ID (P1954)} & P264 & P175 & - & - & - & - & - & \multirow{2}{*}{88.6} \\
             & 84.0 & 93.3 & - & - & -  & -  & -  &  \\
        \midrule
             \multirow{2}{*}{NBIC taxon ID (P8707)} & P171 & - & - & - & - & - & - & \multirow{2}{*}{96.5} \\
             & 96.5 & - & - & - & -  & -  & -  &  \\
        \midrule
             \multirow{2}{*}{Indian Financial System Code (P4635)} & P137 & P17 & - & - & - & - & - & \multirow{2}{*}{50.0} \\
             & 100.0 & 0.0 & - & - & -  & -  & -  &  \\
        \midrule
             \multirow{2}{*}{eBiodiversity ID (P6864)} & P171 & P105 & - & - & - & - & - & \multirow{2}{*}{99.9} \\
             & 99.8 & 100.0 & - & - & -  & -  & -  &  \\
        \midrule
             \multirow{2}{*}{NBN System Key (P3240)} & P171 & P105 & - & - & - & - & - & \multirow{2}{*}{95.6} \\
             & 93.7 & 97.6 & - & - & -  & -  & -  &  \\
        \midrule
             \multirow{2}{*}{New Zealand Organisms Register ID (P2752)} & P171 & P105 & - & - & - & - & - & \multirow{2}{*}{97.6} \\
             & 95.1 & 100.0 & - & - & -  & -  & -  &  \\
        \midrule
             \multirow{2}{*}{Find A Grave memorial ID (P535)} & P734 & P735 & P20 & P19 & - & - & - & \multirow{2}{*}{83.9} \\
             & 85.9 & 77.3 & 85.1 & 87.3 & -  & -  & -  &  \\
        \midrule
             \multirow{2}{*}{ČSFD person ID (P2605)} & P735 & P734 & P19 & - & - & - & - & \multirow{2}{*}{97.4} \\
             & 98.7 & 98.1 & 95.4 & - & -  & -  & -  &  \\
        \midrule
             \multirow{2}{*}{ Ensembl protein ID (P705)} & P527 & P702 & - & - & - & - & - & \multirow{2}{*}{76.2} \\
             & 53.5 & 98.8 & - & - & -  & -  & -  &  \\
        \midrule
             \multirow{2}{*}{ROR ID (P6782)} & P159 & P355 & P17 & P131 & - & - & - & \multirow{2}{*}{81.0} \\
             & 95.5 & 38.0 & 96.8 & 93.8 & -  & -  & -  &  \\
        \midrule
             \multirow{2}{*}{Filmportal ID (P2639)} & P19 & P735 & P57 & P161 & - & - & - & \multirow{2}{*}{71.0} \\
             & 93.6 & 92.4 & 86.5 & 11.5 & -  & -  & -  &  \\
        \midrule
             \multirow{2}{*}{RKDartists ID (P650)} & P20 & P19 & P735 & P937 & - & - & - & \multirow{2}{*}{86.0} \\
             & 89.5 & 93.8 & 91.5 & 69.3 & -  & -  & -  &  \\
        \midrule
             \multirow{2}{*}{NLA Trove ID (P1315)} & P734 & P735 & - & - & - & - & - & \multirow{2}{*}{91.4} \\
             & 93.7 & 89.1 & - & - & -  & -  & -  &  \\
        \midrule
             \multirow{2}{*}{CONOR.SI ID (P1280)} & P734 & P735 & - & - & - & - & - & \multirow{2}{*}{94.1} \\
             & 94.5 & 93.7 & - & - & -  & -  & -  &  \\
        \midrule
             \multirow{2}{*}{Dyntaxa ID (P1939)} & P171 & - & - & - & - & - & - & \multirow{2}{*}{93.4} \\
             & 93.4 & - & - & - & -  & -  & -  &  \\
        \midrule
             \multirow{2}{*}{SIREN number ID (P1616)} & P159 & P1001 & - & - & - & - & - & \multirow{2}{*}{99.0} \\
             & 99.0 & 99.0 & - & - & -  & -  & -  &  \\
        \midrule
             \multirow{2}{*}{Canmore ID (P718)} & P131 & P1343 & P7959 & - & - & - & - & \multirow{2}{*}{99.3} \\
             & 100.0 & 100.0 & 97.9 & - & -  & -  & -  &  \\
        \bottomrule
        \end{tabular}
    }
    
    \caption{Experiment results of supervised learning scenario i.e. 500 examples per property per domain. (PART 2)}
    \label{tab:full_finetune_part_2}
\end{table}

\end{document}